\documentclass[conference]{IEEEtran}
\IEEEoverridecommandlockouts

\usepackage{url}
\usepackage{cite}
\usepackage{amsmath,amssymb,amsfonts}
\usepackage{graphicx}
\usepackage{xcolor}
\usepackage{adjustbox}
\usepackage{siunitx}
\usepackage{makecell}
\usepackage{algorithm}
\usepackage{algpseudocode}
\usepackage{subfig}
\usepackage{booktabs}
\usepackage{gensymb}
\usepackage{colortbl}
\usepackage{enumitem}
\usepackage[font=footnotesize]{caption}

\usepackage{pgfplots}
\usepgfplotslibrary{colorbrewer,statistics,dateplot}
\pgfplotsset{compat=1.18, cycle list/Set1-8} 
\usetikzlibrary{shadows,matrix}

\definecolor{Gray}{gray}{0.925}
\definecolor{blueLine}{RGB}{57,106,177}
\definecolor{blueFill}{RGB}{114,147,203}
\definecolor{redLine}{RGB}{204,37,41}
\definecolor{greenLine}{RGB}{0,250,0}
\definecolor{blackLine}{RGB}{0,0,0}
\definecolor{goldLine}{RGB}{160,82,45}

\graphicspath{{figures/}}
\def\BibTeX{{\rm B\kern-.05em{\sc i\kern-.025em b}\kern-.08em
    T\kern-.1667em\lower.7ex\hbox{E}\kern-.125emX}}

\begin{document}

\title{Data-Efficient Surrogate Modeling for Engineering Design: Ensemble-Free Batch-Mode Deep Active Learning for Regression}

\author{
\IEEEauthorblockN{Sarthak Kapoor}
\IEEEauthorblockA{
Amazon\\
Seattle, WA, USA\\
sarkapoo@amazon.com}
\and
\IEEEauthorblockN{Harsh Vardhan}
\IEEEauthorblockA{Department of Computer Science\\
Vanderbilt University\\
Nashville, TN, USA\\
harsh.vardhan@vanderbilt.edu}
\and
\IEEEauthorblockN{Umesh Timalsina}
\IEEEauthorblockA{Department of 
Computer Science\\
Vanderbilt University\\
Nashville, TN, USA\\
umesh.timalsina@vanderbilt.edu}
\and
\IEEEauthorblockN{Sumit Kumar}
\IEEEauthorblockA{Department of 
Statistics\\
Georgia State University\\
Atlanta, GA, USA\\
skumar18@student.gsu.edu}
\and
\IEEEauthorblockN{Peter Volgyesi}
\IEEEauthorblockA{Department of 
Computer Science\\
Vanderbilt University\\
Nashville, TN, USA\\
peter.volgyesi@vanderbilt.edu}
\and
\IEEEauthorblockN{Janos Sztipanovits}
\IEEEauthorblockA{Department of 
Computer Science\\
Vanderbilt University\\
Nashville, TN, USA\\
jason.sztipanovits@vanderbilt.edu}
}

\maketitle

\begin{abstract}
High-fidelity design evaluation processes such as Computational Fluid Dynamics (CFD) and Finite Element Analysis (FEA) are often replaced with data-driven surrogates to reduce computational cost in engineering design optimization. However, building accurate surrogate models still requires a large number of expensive simulations. To address this challenge, we introduce $\epsilon$-HQS, a scalable active learning (AL) strategy that leverages a student–teacher framework to train deep neural networks (DNNs) efficiently. Unlike Bayesian AL methods, which are computationally demanding with DNNs, $\epsilon$-HQS selectively queries informative samples to reduce labeling cost. Applied to CFD, FEA, and propeller design tasks, our method achieves higher accuracy under fixed labeling cost budgets.
\end{abstract}

\section{Problem Statement}
Let 
\begin{equation}
    y= f(x) , x \in \mathcal{X} \subset \mathbb{R}^n , y \in \mathcal{Y} \subset \mathbb{R}^m 
\end{equation}
be an unknown function for which we aim to construct a surrogate (digital twin) using the training data generated by a high-fidelity simulation process \cite{forrester2008engineering}. Learning such highly complex nonlinear hyper-planes using learning models can assist human designers in finding optimal designs much faster than traditional methods in the optimization loop. Creating a surrogate for complex simulations is challenging due to the curse of dimensionality, where randomized sampling becomes ineffective as data requirements grow exponentially with design space dimensionality \cite{koppen2000curse}. Here, $\mathcal{X}$ represents the unlabeled data set in the design space $\mathcal{DS}$, and $\mathcal{Y}$ represents the solution set evaluated by the simulation process. Accordingly, in the given $\mathcal{DS}$, we view the simulator as a non-linear function that maps the input space to the solution space (\textit{simulator}: $\mathcal{X} \mapsto \mathcal{Y}$), whose behavior we aim to model using a $k$-parameter DNN ($\mathcal{NN}$). For data-driven surrogate modeling, the first step is the generation of training data. 
\begin{equation}
  \mathcal{D}^{train} = \{(x_j^{train} , y_j^{train}) \mid j= 1,2,\ldots,N_{train} \} 
\end{equation}

As numerical simulation processes can be computationally expensive, it is necessary to be strategic during the training data generation process to achieve acceptable accuracy in a minimal number of training data points. This problem can be formalized as:
\begin{align}
\underset{\theta}{\mathrm{argmin}} \quad &\mathbb{E}_{(x,y)}[l(\mathcal{NN}(x,\theta)|\mathcal{D}^{train})] \\
\text{subject to} \quad &\mathcal{D}^{train} \subseteq \mathcal{DS}, \; (x,y)\in \mathcal{D}^{train} \\
&\min \; N_{train} \\
&\mathcal{NN}(x,\theta)=y, \; \forall \, x \in \mathcal{DS} : y\approx y^*
\end{align}

Instead of having a given training data set ($\mathcal{X}$) and its labels ($\mathcal{Y}$), we assume that we don't have any data a priori and the designer must select samples for labeling. To this end, we work in a pool-based setting, where an unlabeled pool ($\mathcal{U}$) containing a set of candidate points in $\mathcal{DS}$ is given for simulation. The heuristic-based research field called Active Learning (AL) tries to choose the most informative samples for evaluation on which to run our simulations.
\begin{equation}
\mathcal{U} = \{x_j \mid j= 1,2,\ldots,N_{pool}\}, \quad \mathcal{U} \subset \mathcal{DS}
\end{equation}
As a subfield of machine learning, AL has been well studied, and if deployed for training a learning model attempts to do so by evaluating the fewest samples
possible. For such a purpose, AL methods rely on an acquisition function ($\mathcal{A}$) which computes a scalar score ($s \in \mathbb{R^+}$) for a trained state of the model and unlabeled pool data ($\mathcal{U}$). 
\begin{equation}
    \mathcal{A}(\mathcal{NN}, \mathcal{U}) : \mathcal{U} \mapsto \mathbb{R}^+ \label{aq}
\end{equation}
The acquisition function ranks the points in $\mathcal{U}$, indirectly measuring the utility of data points for training the surrogate. The unlabeled candidate data point with the maximum score is most appealing for maximizing the model's performance gain in the next iteration of model training, and conversely, points with lower scores are less valuable. Therefore, the design of the acquisition function is crucial to the performance of AL methods.

\section{Approach}
To design the acquisition function $\mathcal{A}$, we employ a neural network referred to as the \textit{teacher} (classifier). 
The surrogate model, termed the \textit{student} network, is the one we seek to train; it learns the simulator’s behavior 
(\textit{student}: $\mathcal{X} \mapsto \mathcal{Y}$), while the teacher network guides sample selection for labeling in the 
next iteration based on the student’s performance and approximates the acquisition function $\mathcal{A}$.

Active learning proceeds iteratively—at each step, both networks are trained. The student is trained using the labeled 
dataset up to the current iteration, $(\mathcal{X}_{\text{labeled}}, \mathcal{Y}_{\text{labeled}})$, which is further divided 
into training and validation subsets. After training on the training portion, we evaluate the student on the validation subset 
and generate a custom dataset by labeling the predicted outputs based on fractional error $\epsilon \in [0,1]$.

\begin{equation}
    \text{Fractional Error} = 
    \frac{|\mathcal{Y}_{\text{labeled}} - \mathcal{Y}_{\text{predicted}}|}%
         {|\mathcal{Y}_{\text{labeled}}|}
\end{equation}

The threshold value $\alpha$ is a user-defined parameter representing the maximum allowable fractional error per test sample. 
A prediction is considered accurate if its error is below $\alpha$; otherwise, it is marked as inaccurate. 
In our experiments, we set $\alpha = 0.05$ (i.e., ±5\% error), which corresponds to a nominal tolerance commonly accepted in engineering design practice. 

Based on this criterion, we assign each sample a binary failure label 
$F: \mathcal{X}_{\text{labeled}} \rightarrow \{0, 1\}$, 
where $F = 0$ indicates acceptable accuracy and $F = 1$ denotes a failure. 
We refer to this custom dataset as the \textit{student test report}. 
This methodology extends naturally to multi-output regression by applying a logical AND operation across all outputs to determine the overall failure indicator $F$.

\begin{equation}
\begin{aligned}
    \mathcal{D}_{\text{labeled}}
        &= \left\{ (x_i, F_i) \;\middle|\; x_i \in \mathcal{X}_{\text{labeled}} \right\}, \\[4pt]
    \text{where} \quad
        F_i &= \mathbb{I}\!\left(\frac{|y_i - y_i^*|}{|y_i^*|} \geq \alpha\right)
\end{aligned}
\end{equation}

\begin{figure*}[ht!]
    \centering
    \includegraphics[width=15.5 cm]{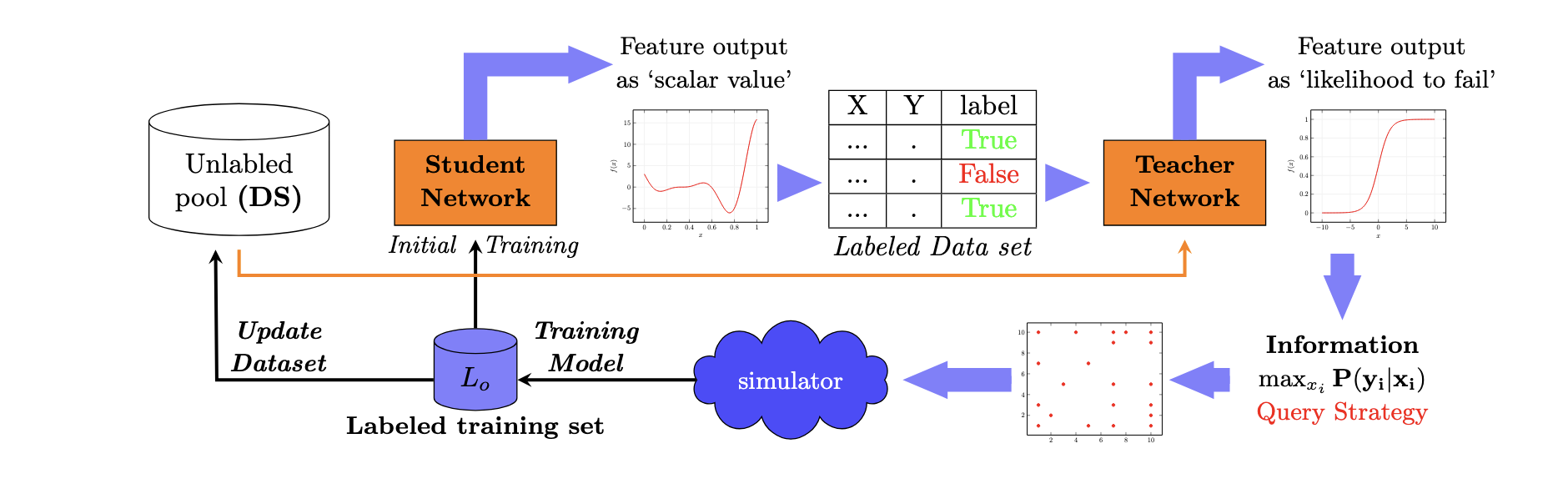}
    \caption{Student Teacher Architecture: The Process}
    \label{fig:st}
\end{figure*}

Using the \textit{student test report}, the teacher network is trained to learn a failure probability distribution of the student over the entire design space ($\mathcal{DS}$). The teacher network models the failure probability and guides the sampling in the next iteration toward regions in $\mathcal{DS}$ where the student network has poor performance/high likelihood of failure. At each iteration of learning, we retrain both the student and teacher networks on newly evaluated samples until we exhaust our total budget (refer to Figure~\ref{fig:st}).
Since the teacher performs a classification task, it uses a smaller neural network than the student, as approximating a decision boundary is simpler than learning a complex regression manifold. Leveraging GPU acceleration, the teacher can efficiently evaluate thousands of samples with minimal computational cost compared to running the full simulation. Using the predicted failure probabilities, we can identify the samples $x \in \mathcal{DS}$ where the student is most likely to fail in each iteration.

\begin{equation}
    \mathcal{S}^* = \underset{x \in \mathcal{U}}{\arg\max} \; f_t(x)
\end{equation}

\subsection{Batching}
Classical active learning (AL) methods often select a single sample per iteration, leading to minimal updates to the training set. This has two drawbacks: (1) retraining the network for each new sample is time-consuming and negates the efficiency gains of AL, and (2) frequent retraining on nearly identical data can cause overfitting \cite{ren2021survey}. To overcome this, we adopt batch-mode active learning by selecting a set of candidate unlabeled samples $\mathcal{B} =\{x_1 ,x_2 ,...,x_b\} \subseteq \mathcal{U}$. Using the acquisition function $\mathcal{A}$ provided by the trained teacher, we aim to select an optimal batch $\mathcal{B}^*= \{x_1^* ,x_2^* ,...,x_b^*\}$ at each iteration, formulated as:
\begin{equation}
     \mathcal{B}^* = \underset{\mathcal{B}\subseteq \mathcal{U}} {\mathrm{argmax}} \; \mathcal{P}(\mathcal{B},f_t(x)) 
\end{equation}
Here, $\mathcal{P}$ is a policy, algorithm, or heuristic that chooses a batch of samples. In this work, we explore the following policies for batch selection, drawn from current research in Active Learning (AL): 
\begin{enumerate}
    \item Top-b : It is a greedy strategy, here $b$ is the batch size at each iteration. In this approach, after evaluating the failure probability score on all unlabeled samples $\mathcal{U}$, top-$b$ samples are selected that have the maximum probability of failure. 
    \item Diverse Batched Active Learning (DBAL) : The general approach to impart batched collective maximum information in AL literature is to include diversity in the sample selection. Wei et al. \cite{wei2015submodularity} include diversity by formulating a submodular function on the distances between samples and selecting a batch of unlabeled samples, which optimizes the submodular function.
    \item Batched Random: This approach samples a batch of random samples $\mathcal{B}$ from the unlabeled pool $\mathcal{U}$ and label it at each iteration of training.
    \item Epsilon($\epsilon$)-weighted Hybrid Query Strategy ($\epsilon$-HQS) : Proposed in this work and explained below. 
\end{enumerate}

\subsection{\textbf{Epsilon($\epsilon$)-weighted Hybrid Query Strategy}}
We propose a new batched sampling policy, $\epsilon$-HQS, inspired by the exploration–exploitation trade-off in reinforcement learning \cite{sutton2018reinforcement, vardhan2022deepal}. The complete pseudo-code for this approach is given in Algorithm \ref{alg:ewhqs}.

\begin{algorithm}
\caption{$\epsilon$-HQS}\label{alg:ewhqs}
\begin{algorithmic}[1]
\State Require: {student\; network $f_s(,\theta)$; teacher network $f_t(x,w)$; unlabeled sample pool $\mathcal{U}$; number of initial samples $M$; number of AL iterations $T$; batch size $b$, acceptable accuracy $\alpha$, hyper-parameter $\epsilon$
\State Create labeled data: $S \leftarrow M$ examples drawn uniformly randomly from $\mathcal{U}$ and label it.
\State Train the student model to get $\theta_1$ on $S$ by minimizing ${E}_{S}[l_{MAE}(f_s(x;\theta),y)]$
\State Create labeled data: $\mathcal{D}_{labeled} = \{x_i,\mathbb{I}(y_i-y_i^*) \geq \alpha \times y_i\} \; \forall \;x_i \in S$
\State Train the teacher model to get $w_1$ on $\mathcal{D}_{labeled}$ by minimizing ${E}_{\mathcal{D}_{labeled}}[l_{CE}(f_s(x;w),y)]$

\For{$t=1,2,...,T:$} 
\State For all examples $x$ in $\mathcal{U}  \setminus S$:
 \begin{enumerate}
  \item Compute its failure probability $p(x)= f_t(x)$
  \item $ X_f = \{x \;|\; p(x) \geq 0.5\}$ \Comment{ samples with high failure probability}
  \item $X_{sel} = Choose\;(\epsilon \times b)\; number\; of \;samples\; uniformly\; randomly\; from\; X_f$. 
  \item $S \leftarrow S+X_{sel}$
  \item $X_{rest}= Choose\;((1-\epsilon) \times b)\; number\; of \;samples\; uniformly\; randomly\; from\; \mathcal{U} \setminus X_{sel}$ 
  \item $S \leftarrow S+X_{rest}$
 \end{enumerate}
\State $S \leftarrow S \bigcup S_t $
\State Train the student model to get $\theta_{t+1}$ on $S$ by minimizing ${E}_{S}[l_{MAE}(f_s(x;\theta_{t}),y)]$
\State Create labeled dataset: $\mathcal{D}_{labeled} = \{x_i,\mathbb{I}(y_i-y_i^*) \geq \alpha \times y_i\} \; \forall \;x_i \in S$
\State Train the teacher model to get $w_{t+1}$ on $S$ by minimizing ${E}_{\mathcal{D}_{labeled}}[l_{CE}(f_s(x;w_t),y)]$
\EndFor
\State \textbf{return} surrogate student model and its weights $\theta_{T+1}$
}\end{algorithmic}
\end{algorithm}

In our student–teacher framework, the teacher’s role is to estimate the surrogate model’s performance across the design space by assigning a failure probability to each sample ($f: x \mapsto [0,1],; \forall; x \in \mathcal{DS}$). At each iteration of teacher training, we want to estimate the following:
 \begin{equation}
     f^*(x)= \mathbb{P}(f_s(x)\geq \alpha \cdot y^*), \quad x \in \mathcal{DS}, \; y^*= \text{true label}
 \end{equation}
If $f^*$ is the true failure probability of the student model in entire design space ($\mathcal{DS}$). However, since at any iteration of training, we do not have true labels ($y^*$) for all unlabeled points in $\mathcal{U}$. In such a case, the goal of a teacher is to learn an approximation $f \approx f^*$. As at each iteration, $\mathcal{D}_{labeled}$ has limited information about the design space $\mathcal{DS}$ (a maximum up to training and testing data), we want this approximation to be as accurate as possible.    
 \begin{equation}
     f_t(x)= \mathbb{P}(f_s(x,w)|\mathcal{D}_{\text{labeled}}), \quad \forall\; x \in \mathcal{DS}
 \end{equation}
In such a case, we can only approximate $f^*$ with some relative accuracy ($\rho$ approximation  of $f^*$).  It results in the biased estimation of the performance of the student model by the teacher model. Estimation bound on $\rho$ is an open research question, but we can increase the robustness of the teacher network and can reduce the sensitivity to mismatch between the distribution of $f_t$ and $f^*$. For this purpose, we introduce a two-step process: first, we filter all samples with teacher's prediction (likelihood to fail) more than a threshold value (samples with high probability of failure), and second, we introduce a belief weight-age on the teacher network. For this end, we introduce two hyperparameters- threshold and $\epsilon$. Since the output layer of the teacher network is a sigmoid function, we kept the standard threshold of $0.5$ for all experiments (i.e., select all $x$ if $f_t(x)\geq 0.5$; $\forall \;x \in \mathcal{DS}$). $\epsilon$ is a belief factor that we assign to the teacher network.  $\epsilon$ is a scalar real value ranging between $0$ (indicates complete disbelief in the teacher network's estimate) and $1$ (indicates a complete belief in teacher's estimation) ($\epsilon \in [0,1]$).
 
 $\epsilon-greedy$ based policy is inspired by RL \cite{sutton2018reinforcement} literature, the $\epsilon$ factor controls the balance between exploration and exploitation. Since, the teacher network does not know what it does not know - the exploration versus exploitation dilemma exists in this situation similar to RL, i.e., the search for a balance between exploring the design space to find regions where we have not explored  while exploiting the knowledge gained so far by the teacher network. The fixed value of $\epsilon$ expresses linear belief in the teacher prediction. Another famous approach is to let $\epsilon$ go to one with a certain rate to increase our belief in the teacher with more and more labeled data available for its training in later iterations. It turns out that at a rate of $\frac{1}{T}$($T=\text{number of training iterations}$) proved to have a logarithmic bound on the regret (maximum gain) \cite{auer2002finite}.  $\epsilon-greedy$ rule \cite{sutton2018reinforcement}  is a simple and well-known policy for the bandit problem.
At each iteration of active sampling, this policy selects $\epsilon \times b$ number of samples from the filtered unlabeled data (samples that have a likelihood to fail more than the threshold) and the rest of the samples $(1-\epsilon)\times b$ is selected uniformly randomly from leftover unlabeled samples. The complete pseudo-code for this approach is given in the algorithm \ref{alg:ewhqs}.

\section{Experiments and Results}
In our experiments, we follow a pool-based Active Learning (AL) setup, starting with a fixed pool of unlabeled data ($\mathcal{U}$). At each iteration, a batch of samples is selected for labeling based on a query strategy. The selected labeled data is added to the training set ($\mathcal{D}_{\text{train}}$) to iteratively improve the student model. This process repeats for $T$ iterations. Model performance is evaluated using the remaining unlabeled data, and results are averaged over multiple independent runs to ensure robustness. We also compare simulation labeling time and surrogate prediction time to assess computational efficiency. We evaluate surrogate performance using the fraction of test predictions within ±5\% of the ground truth, as defined by the accuracy metric in the following equation:
\begin{equation}
    Accuracy =\frac{\sum_{i=1}^{N_{test}} \mathbb{I}(|y_i-y_i^*| \leq 0.05*|y_i^*|)}{N_{test}}
\end{equation}

\begin{figure}[htbp]
  \centering
  \subfloat[Propeller Domain]{\includegraphics[clip, trim=0.1cm 0.1cm 0.1cm 0.1cm, width=0.45\textwidth]{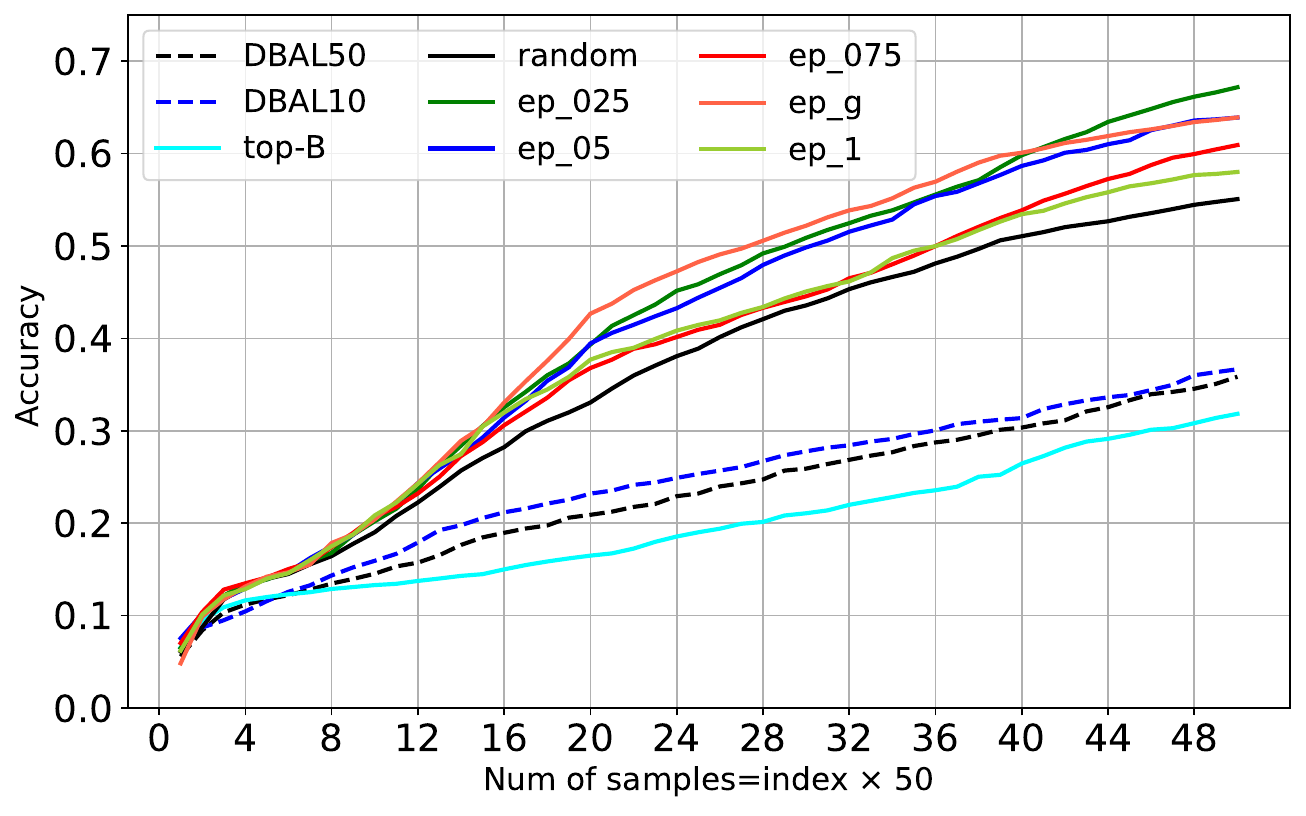}\label{fig:prop}}
  \hfil
  \subfloat[FEA domain]{\includegraphics[clip, trim=0.1cm 0.1cm 0.1cm 0.1cm, width=0.45\textwidth]{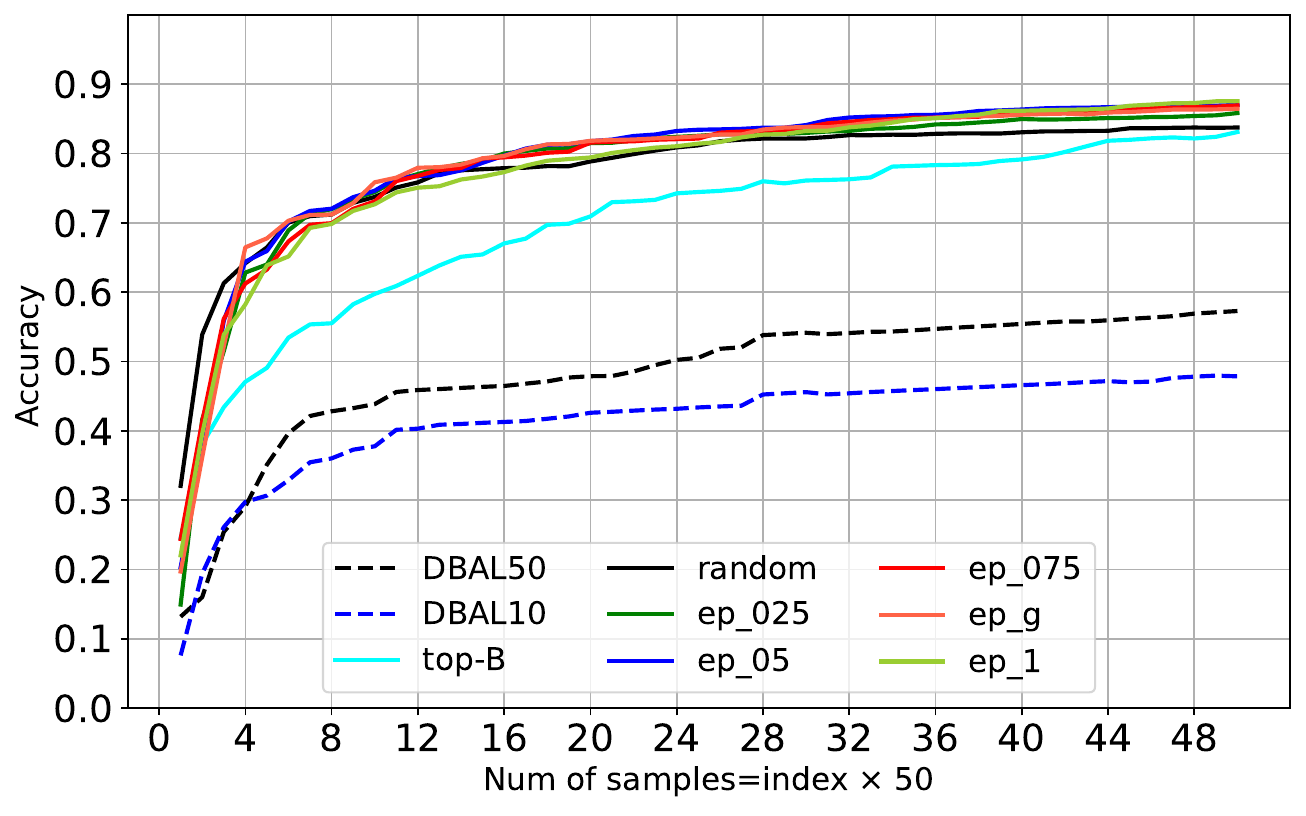}\label{fig:fea}}
  \hfil
  \subfloat[CFD domain]{\includegraphics[clip, trim=0.1cm 0.1cm 0.1cm 0.1cm, width=0.45\textwidth]{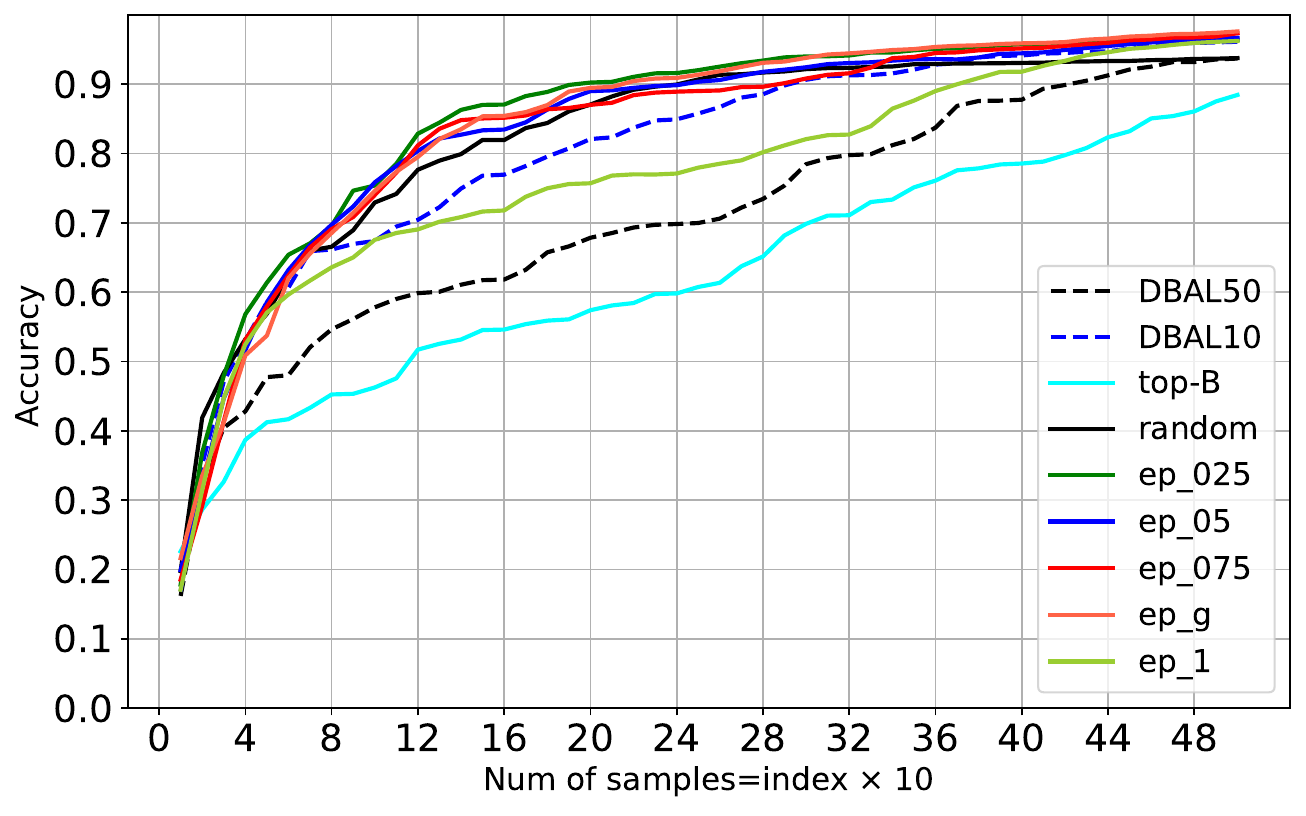}\label{fig:cfd}}
  \caption{The comparison of expected/mean test accuracy of trained surrogate in different domain using all proposed approaches at the different iterations of training. DBAL50 (Diverse Batch Active Learning with $\beta$=50), DBAL10 (Diverse Batch Active Learning with $\beta$=10), random (batch uniformly random), ep\_025 ($\epsilon-HQS$ with constant $\epsilon$=0.25), ep\_05 ($\epsilon-HQS$ with constant $\epsilon$=0.5), ep\_075 ($\epsilon-HQS$ with constant $\epsilon$=0.75), ep\_1 ($\epsilon-HQS$ with constant $\epsilon$=1.0), ep\_greedy ($\epsilon-HQS$ with logarithmic increasing $\epsilon$).}
  \label{fig:main}
\end{figure}

\begin{table*}
\centering
\normalsize
\captionsetup{justification=centering,format=plain,font=small, labelfont=bf}
\begin{tabular}{lccc} 
\toprule
\textbf{Sampling strategy} & \hfil \textbf{Accuracy (CFD)}  & \hfil \textbf{Accuracy (FEA)}  & \hfil \textbf{Accuracy (Propeller)}\\
\midrule
Random & $93.78\%$ & $83.76\%$  & $55.07\%$  \\
top-b  & $88.46\%$  & $83.11\%$  & $31.83\%$\\
DBAL-10 & $93.71\%$ & $57.28\%$ & $35.84\%$  \\
DBAL-50  & $96.13\%$ & $47.96\%$  & $36.66\%$  \\
$\epsilon$--greedy  & $97.62\%$  & $86.47\%$  & $63.9\%$  \\
$\epsilon=0.25$  & $96.43\%$  & $85.85\%$ & $67.18\%$ \\
$\epsilon=0.50$  & $96.73\%$ & $87.01\%$ & $63.9\%$\\
$\epsilon=0.75$  & $97.37\%$ & $86.96\%$ & $60.92\%$  \\
$\epsilon=1.0$  & $96.25\%$ & $87.56\%$ & $58.01\%$ \\
\bottomrule
\end{tabular}
\vspace{0mm}
\caption{Accuracy table in different design domain after exhaustion of budget.(CFD: Iterations=50, budget/iteration=10, Total budget = 500) ; (FEA \& Propeller : Iterations=50,budget/iteration=50 ; Total budget = 2500)}
\label{tab:cfd_res}
\end{table*}

For the FEA domain, the goal is to predict von Mises stress, which is a key metric for evaluating the structural integrity of subsea pressure vessels housing sensitive components under external seawater pressure. Vardhan et al. \cite{vardhan2022deep} studied a cylindrical vessel made of Aluminum Alloy Al-6061-T6 with hemispherical end-caps to ensure efficient pressure distribution. Accurate Finite Element Analysis (FEA) was achieved through iterative mesh refinement until solution convergence. The design space is defined by the design parameters and sea depth, which determines the external pressure. The applied crush pressure is estimated using hydrostatic pressure theory with seawater density of 1027 kg/m³. The design space for this experiment comprises four variables: sea depth ($D_{sea}$), vessel length ($L$), external hull diameter ($d$), and hull thickness ($t$), i.e., $\mathcal{DS}={D_{sea},L,d,t}$ where $\mathcal{DS} \in \mathbb{R}^4$ maps to von Mises stress, indicating whether the structure can resist external hydrostatic pressure.

For the propeller design domain, the goal is to predict propeller efficiency, which is influenced by geometric features (i.e., number of blades, diameter, pitch/chord distributions) and physical parameters (e.g., thrust and power coefficients, advance ratio). Given operational constraints such as required thrust and RPM, the design objective is to optimize geometry for maximum efficiency. Data generation employs OpenProp \cite{epps2009openprop} as the propeller design tool. For data generation, we utilized the same design space as Vardhan et al. \cite{vardhan2021machine}. Since the evaluation of propeller design using the OpenProp simulation tool is relatively less computationally expensive than FEA and CFD, and given the higher dimensionality of the problem (14 dimensions), we generated approximately 200,000 valid datapoints as pool data.

For the CFD domain, the goal is to learn the drag resistance of an Unmanned Underwater Vehicle (UUV) to minimize drag. For data generation, we adopt the widely used Myring hull profile \cite{myring1976theoretical}, which consists of a nose, cylindrical body, and tail. The nose and tail shapes are parameterized using analytical equations to define the design space. To estimate drag for hull designs, we employ Computational Fluid Dynamics (CFD) using the Reynolds-Averaged Navier-Stokes (RANS) equations \cite{sagaut2013multiscale} with the $k$-$\omega$ SST turbulence model \cite{menter1992improved}. Simulations are performed using the open-source tool OpenFOAM \cite{jasak2007openfoam}. The design space comprises six parameters: diameter of middle section ($d$), length of nose section ($a$), length of middle section ($b$), length of tail section ($c$), index of nose shape ($n$), and tail semi-angle ($\theta$), i.e., $\mathcal{DS}={d,a,b,c,n,\theta}$ where $\mathcal{DS} \in \mathbb{R}^6$ maps to the drag value.

Figures \ref{fig:cfd}, \ref{fig:fea}, and \ref{fig:prop} show the accuracy of trained surrogates/digital twins at different training iterations in each domain. Table \ref{tab:cfd_res} shows the final accuracy metrics using different acquisition function strategies. Results demonstrate that $\epsilon$-HQS consistently outperforms other acquisition function strategies in all experiments. Adopting this strategy can significantly reduce data labeling time due to its sample efficiency, saving hours to days of training time compared to other methods.

\section{Related work}
Surrogate modeling in engineering design involving regression problems is a widely studied and prevalent approach when computationally complex engineering domains are involved in the design process \cite{forrester2008engineering,simpson2001metamodels}. Historically, surrogate modeling has primarily employed learning models such as kriging \cite{forrester2008engineering}, Gaussian processes \cite{rasmussen2003gaussian}, support vector machines (SVM) \cite{pisner2020support}, and random forests \cite{breiman2001random}. However, these methods face scalability challenges with respect to both data size and design space dimensionality, as their computational complexity becomes intractable for high-dimensional problems \cite{palar2019use}. Due to these limitations, surrogate modeling approaches have traditionally been restricted to problems with limited design spaces, constraining their utility in design optimization.

\section{Conclusions}
This work presents a scalable deep active learning method, $\epsilon$-HQS, for surrogate modeling in regression tasks. Experiment results show that $\epsilon$-HQS consistently outperforms other methods, with its advantage growing in higher-dimensional problems.
\bibliographystyle{IEEEtran}
\bibliography{bibliography}

\end{document}